\documentclass{article}

\usepackage{microtype}
\usepackage{graphicx}
\usepackage{subfigure}
\usepackage{booktabs} % for professional tables

\usepackage{hyperref}

\usepackage[accepted]{icml2023/icml2023}

\usepackage{amsmath}
\usepackage{amssymb}
\usepackage{mathtools}
\usepackage{amsthm}

\usepackage[capitalize,noabbrev]{cleveref}

\theoremstyle{plain}

\theoremstyle{definition}

\theoremstyle{remark}

\newcommand{\internshipCommand}{\textsuperscript{$\dagger$}Work done while an intern at Intel}

\usepackage[textsize=tiny]{todonotes}

\icmltitlerunning{XLDA: LDA for Scaling CL to Extreme Classification at the Edge}

\begin{document}

\twocolumn[
  \icmltitle{XLDA: Linear Discriminant Analysis for Scaling Continual Learning to Extreme Classification at the Edge}

  % It is OKAY to include author information, even for blind
  % submissions: the style file will automatically remove it for you
  % unless you've provided the [accepted] option to the icml2023
  % package.

  % List of affiliations: The first argument should be a (short)
  % identifier you will use later to specify author affiliations
  % Academic affiliations should list Department, University, City, Region, Country
  % Industry affiliations should list Company, City, Region, Country

  % You can specify symbols, otherwise they are numbered in order.
  % Ideally, you should not use this facility. Affiliations will be numbered
  % in order of appearance and this is the preferred way.
  \icmlsetsymbol{equal}{*}
  \icmlsetsymbol{intern}{$\dagger$}

  \begin{icmlauthorlist}
    \icmlauthor{Karan Shah}{equal,intel}
    \icmlauthor{Vishruth Veerendranath}{equal,intern,pes}
    \icmlauthor{Anushka Hebbar}{equal,intern,pes}
    \icmlauthor{Raghavendra Bhat}{intel}
  \end{icmlauthorlist}

  \icmlaffiliation{intel}{Intel Corporation}
  \icmlaffiliation{pes}{PES University, Bangalore, India}
  % \icmlaffiliation{intern}{Work done while an intern at Intel Corporation}

  \icmlcorrespondingauthor{Karan Shah}{karan.shah@intel.com}
  \icmlcorrespondingauthor{Vishruth Veerendranath}{vishruthnath@gmail.com}

  % You may provide any keywords that you
  % find helpful for describing your paper; these are used to populate
  % the "keywords" metadata in the PDF but will not be shown in the document
  \icmlkeywords{Continual Learning,Extreme Classification,Edge Optimizations,LDA}

  \vskip 0.3in
]

% this must go after the closing bracket ] following \twocolumn[ ...

% This command actually creates the footnote in the first column
% listing the affiliations and the copyright notice.
% The command takes one argument, which is text to display at the start of the footnote.
% The \icmlEqualContribution command is standard text for equal contribution.
% Remove it (just {}) if you do not need this facility.

%\printAffiliationsAndNotice{}  % leave blank if no need to mention equal contribution
\printAffiliationsAndNotice{\icmlEqualContribution \internshipCommand} % otherwise use the standard text.

\begin{abstract}
  % Linear Discriminant Analysis (LDA) outperforms most Class-incremental Learning techniques by leveraging fixed feature extractors and learning a simple per-class linear classifier, in a streaming fashion.
  % While Streaming LDA has demonstrated success in addressing CL scenarios with up to 1000 classes; in certain edge deployments like Retail, class counts exceed well beyond this range. 
  % Training `extreme' class count classifiers comes with a quadratic increase in compute requirement. On the other hand, LDA being a gradient-free algorithm, cannot be used to train entire model end-to-end. 
  % Our contribution in this work is two-fold: First, we extend evaluation of LDA for upto 50k classes, showing that LDA classifiers are on par with fully-connected layers by accuracy, 
  % while describing a comprehensive set of algorithmic and computational optimizations that lead to convergence rates that are up to nX faster than fully-connected layers. 
  % Second, we show that under certain assumptions, a fully-connected layer can be converted to an LDA classifier, and vice versa, without significant loss in accuracy, eliminating the need for finetuning or retraining at the edge.
  % We experiment on datasets with upto 200k classes.

  Streaming Linear Discriminant Analysis (LDA) while proven in Class-incremental Learning deployments at the edge with limited classes (upto 1000), has not been proven for deployment in extreme classification scenarios.
  In this paper, we present: (a) XLDA, a framework for Class-IL in edge deployment where LDA classifier is proven to be equivalent to FC layer including in extreme classification scenarios,
  and (b) optimizations to enable XLDA-based training and inference for edge deployment where there is a constraint on available compute resources.
  We show upto \textbf{42$\times$} speed up using a batched training approach and upto \textbf{5$\times$} inference speedup with nearest neighbor search on extreme datasets like AliProducts (50k classes) and Google Landmarks V2 (81k classes).
  % Our contribution in this work is two-fold: First, we extend evaluation of LDA for upto 50k classes, showing that LDA classifiers are on par with fully-connected layers by accuracy, while describing a comprehensive set of algorithmic and computational optimizations that lead to convergence rates that are up to nX faster than fully-connected layers. Second, we show that under certain assumptions, a fully-connected layer can be converted to an LDA classifier, and vice versa, without significant loss in accuracy, eliminating the need for finetuning or retraining at the edge.

\end{abstract}

\section{Introduction}
Training models using all data previously available as well as data that arrives in small batches, comes with compute constraints at the edge. It mandates the need for models that can adapt to incrementally arriving data, without stressing local compute. In classification, this is commonly known as the \textbf{Class-incremental Learning} objective.

Standard DNN-based classifiers when trained incrementally, suffer from catastrophic forgetting. Previous literature proposes various ways of dealing with forgetting; using external memory \cite{rebuffi2017icarl, lopez2017gradient, van2020brain}, by constraining parameter \cite{kirkpatrick2017overcoming, zenke2017continual} or activation drift \cite{li2017learning}.

By far, most successful methods operate on storing or generating data for rehearsal, which requires models to be finetuned on the same data again, increasing the compute footprint. This gets exacerbated when the number of classes grows significantly over time, as is the case in many use-cases at the edge like retail, healthcare and education.

\begin{figure}
  \centering
  \includegraphics[width=\columnwidth]{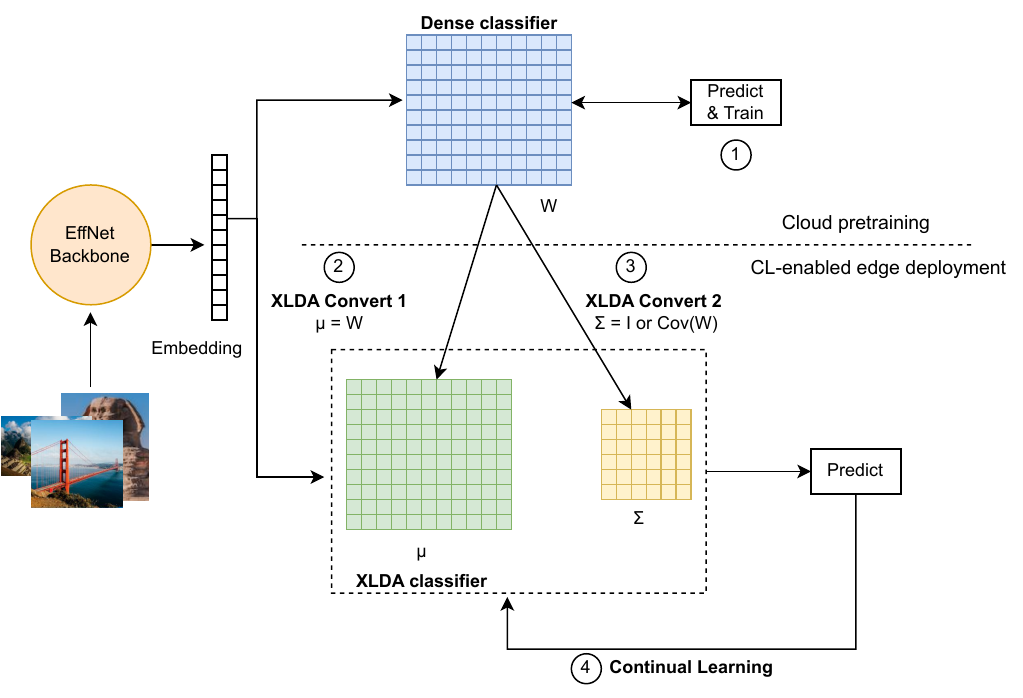}
  \caption{Deployment scenario for FC-XLDA conversion }
  \label{fig:base_init}
\end{figure}

Handling classification with a large number of classes has been an active area of research, referred to as \emph{Extreme Multi-label Classification} (XML) \cite{Bhatia16}. These methods \cite{siblini2018craftml,dahiya2021deepxml,yu2022pecos,dahiya2023ngame} have been used for search and retrieval on text data consisting of millions of labels. In the image domain, this has been explored in a few recognition tasks \cite{schroff2015facenet,deng2019arcface,weyand2020landmarks,ypsilantis2021met,song2020knnsoftmax}, but not in Class-IL settings.

In this work, we focus on the task-free Class-IL setting \cite{aljundi2019task}, without replay or rehearsal, which is pragmatic in real-world deployments. Since our focus here is on class counts exceeding 1k, we assume feature extractors to be pretrained on a relevant larger subset on the cloud. At the edge, the problem reduces to learning a classifier incrementally on the features of new data.

Linear Discriminant Analysis (LDA) \cite{fisher1936linear} has been popularly used for dimensionality reduction, and it lends itself very well to Class-IL \cite{pang2005incremental}, as a classifier on top of pretrained model features. \citet{hayes2020lifelong} demonstrated this use of LDA, and achieved competitive or better accuracy against all other techniques on datasets up to 1k classes, even when the data was presented only once, in a streaming manner.

In this paper, we present E\textbf{X}treme \textbf{LDA} \textbf{(XLDA)}, a framework of optimizations that enables scaling LDA-based classifiers to extreme multiclass settings. Our contributions are as follows:
\begin{itemize}
  \item \textbf{Class-IL beyond 1k classes:} We extend LDA benchmarks on Class-IL up to 81k classes, showing that it can achieve competitive or on-par accuracy of a fully-connected (FC) layer trained until convergence, in Section \ref{sec:equivalence}. Benchmarks are covered in Section \ref{sec:experiments}.
  \item \textbf{Optimizing training and inference:} LDA can be used as a drop-in replacement of an FC layer with $O(1)$ time complexity training per sample. We parallelize this for batches in Section \ref{sec:batch_parallel}, which results in a worst-case \textbf{42$\times$} speedup in training versus FC layers for 50k classes. We optimize inference for upto \textbf{5$\times$} speedup using nearest neighbor search in Section \ref{sec:inference_optim_method}. The experiments for these are detailed in Section \ref{sec:experiments}.
  \item \textbf{Translating between LDA and FC:} We empirically show that an FC layer can be converted to an LDA layer, while maintaining accuracy, to enable Class-IL at the edge, under certain assumptions, in Section \ref{sec:conversion}.
\end{itemize}

\section{Optimizations in XLDA}
Background on SLDA has been included in Appendix \ref{app:SLDA}.

\subsection{Conversion from a Fully-Connected layer} \label{sec:base_init}
In Deep SLDA \cite{hayes2020lifelong} authors train two components: \emph{feature extractor} and
the \emph{LDA classifier} separately. We posit that a model trained end-to-end using an FC layer (with parameters $W$ and $b$), can be converted to an LDA classifier (with parameters $\mu$ and shared covariance $\Sigma_s$), assuming samples are normally distributed with identity covariance across all classes.

% First, the CNN backbone is trained with a fully connected classification head on the initialization data, and then the embeddings of all images are extracted by performing a forward pass through the frozen backbone.
% Secondly, using these embeddings and the class labels, the mean vector for each class and the covariance matrix is updated one sample at a time.

% This duplication of effort in doing two passes over the dataset one sample at-a-time, leads to the slowdown of training process.
% To reduce the training time and computation, we propose initializing SLDA's parameters
% with information already present in the parameters of the fully connected layer used while training the CNN.

Let $W$ be the weights of the last fully connected layer used to train the model. In order to replace this with an LDA classifier of equivalent performance, we initialize the covariance matrix $\Sigma$ to either of the two options below
\begin{equation}
  \begin{aligned}
    \Sigma & = \mathrm{I}      \\
    \Sigma & = \mathrm{Cov}(W)
  \end{aligned}
\end{equation}
where, $I$ $\in \mathbb{R}^{d \times d}$ is an identity matrix, and $Cov(W)$ is the covariance of $W$.
The first initialization $\Sigma = \mathrm{I}$, follows our assumption of decorrelated dimensions in high-dimensional spaces if the classes are easily distinguishable. The second initialization $\Sigma = \mathrm{Cov}(W)$, is empirically observed to handle the "extreme" cases where due to a large number of classes and inter-class confusion, the dimensions of the embeddings would be correlated. Here, we hypothesize that the correlation in embeddings would be captured implicitly in the correlation between the weight matrix's dimensions.

We then simply initialize the means matrix as $\mu = W$. This follows from SLDA's Eqn. \ref{eq:mu_w_flip} (detailed in Appendix \ref{app:inference})  and
the previous initialization of $\Sigma = \mathrm{I}$.

% Let $W$ be the weights of fully connected layer after the CNN has been fit to data.
% For $\mu$, we hypothesize that the cross-entropy minimization objective implicitly learns the mean embedding for each class.
% We then simply initialize the means matrix $\mu$ as $\mu = W$, as they are both of shape $(N \times d)$. 

% The covariance matrix is initialized to $\Sigma = I$ where $I$ denotes an identity matrix of shape $(d \times d)$.

% If both these conditions are satisfied, the covariance matrix would reduce down to an identity matrix.

% After the base initialization step, $\Sigma$ can be kept plastic i.e it gets updated with each sample seen in the continual learning setting.
Performing the base initialization in this way also helps transfer pre-trained classifier weights to an SLDA classifier
that can be deployed at the edge with CL-capabilities with no retraining effort as seen in Fig. \ref{fig:base_init}.
The assumption of using information in dense layer for SLDA is further validated by the equivalence of LDA and fully connected layer shown in Section \ref{sec:equivalence}.

\subsection{Batch Parallel Training} \label{sec:batch_parallel}
% In the case of discriminative classifiers, training time is generally bottlenecked by the number of classes in the dataset. In the case of a fully connected layer,
% training is $O(K)$, where $K$ is the number of classes, as both the update of the weight matrix $W$ during backpropagation and the computation of final softmax layer
% scale linearly with the number of classes. This process becomes an extremely expensive affair in extreme learning settings. 

LDA-based classifiers can theoretically be trained in constant time with respect to the number of classes, as the process only involves
updating for every sample, the corresponding class’s mean vector $\mu_k$ and the shared covariance $\Sigma$. This makes the time complexity of training SLDA $O(1)$ with respect to the number of classes
and as a result, highly optimized for training in extreme multiclass settings.
% Details of updates during conventional SLDA training are included in Appendix \ref{app:SLDA_train}.
Details of training updates detailed by \citet{hayes2020lifelong} are included in Appendix \ref{app:SLDA_train}.

One of the upsides of a fully connected layer is its ability to handle vectorized batch parallel training efficiently on specialized hardware, which is much faster than per-sample training strategies.
Inspired by chunk updates described in \citet{pang2005incremental}, we vectorize the training process of Deep SLDA \cite{hayes2020lifelong} and incorporate batch-parallelism which updates the mean and covariance parameters just once for every batch of samples.
The method detailed in Algorithm \ref{alg:batch_parallel},
allows for better utilization of specialized hardware as well as a significant speedup in training times.
% Batch training updates model parameters only once per batch
% and can be made efficient by utilizing vectorization which parallelizes computation on the entire batch on hardware. SLDA on the other hand, updates two parameters
% for every sample it encounters – the corresponding class mean vector and the shared covariance matrix.  
% While SLDA provides far faster training compared to a fully connected layer under extreme multiclass settings, incorporating batch parallelism
% and providing a method of updating the mean and covariance parameters just once for every batch of samples as shown in Algorithm \ref{alg:batch_parallel}, could further optimize training times.
% – which would also be an asset in extreme learning scenarios.

\begin{algorithm}[tb]
  \caption{Batch Parallel Training in XLDA}
  \label{alg:batch_parallel}
  \begin{algorithmic}
    \STATE
    \texttt{m}: batch size, \texttt{n}: number of classes \\
    \texttt{d}: length of a feature embedding \\
    \STATE {\bfseries Input:} Batch of data (\textbf{X}, \textbf{y}) where $\textbf{X} \in \mathbb{R}^{m \times d}$, $\textbf{y} \in \mathbb{R}^{m}$
    \STATE Initialize means $\boldsymbol{\mu} \in \mathbb{R}^{n \times d}$ and counts $\textbf{C} \in \mathbb{R}^{n}$ to zero
    \FOR{every batch of samples (\textbf{X}, \textbf{y})}
    \STATE Compute a dictionary of unique labels and their counts in \textbf{y} where $l$ is a unique label and $s$ is its count in \textbf{y}
    \STATE U: dict($l:s$) $\gets$ UNIQUE(\textbf{y})
    \STATE \textbf{parallel for} label $i$ in \textsc{U}\texttt{.keys} do
    \STATE \hspace{8mm} $\textbf{C}[i] \gets \textbf{C}[i] + \textsc{U}[i]$
    \STATE \hspace{8mm} $\boldsymbol{\mu}[i] \gets \boldsymbol{\mu}[i] + \frac{\textbf{X}_i - \boldsymbol{\mu}[i]}{\textbf{C}[i]}$
    \STATE \textbf{end for}
    \ENDFOR
  \end{algorithmic}
\end{algorithm}

% In order to support updating the mean
% parameter just once after the introduction of an entire batch of samples, we derive the rule to update the mean vectors corresponding to 
% every class. 

% \begin{equation}
%   \begin{aligned}
%     \boldsymbol{\mu}_{(k=y,t+\tau)} \gets \frac{c_{(k=y,t)}\boldsymbol{\mu}_{(k=y,t)} + \displaystyle\sum_{} \mathbf{z}_{t}}{ c_{(k=y,t)} + a} \\
%     c_{(k=y,t+1)} = c_{(k=y,t)} + a
%   \end{aligned}
% \end{equation}

% This can also be stated as an addition update to the existing mean vector as,

% \begin{equation}
%   \begin{aligned}
%     \boldsymbol{\mu}_{(k=y,t+\tau)} \gets \boldsymbol{\mu}_{(k=y,t)} + \frac{\displaystyle\sum_{i=0}^{i=\tau-1}\mathbf{z}_{t+i} - \tau \boldsymbol{\mu}_{(k=y,t)}}{ c_{(k=y,t)} + \tau} \\
%     c_{(k=y,t+1)} = c_{(k=y,t)} + \tau
%   \end{aligned}
% \end{equation}

% Since this rule only depends on the values of the batch's samples, and the mean and count variables at timestep $t$, the mean parameter can be updated
% at once per batch instead of per sample.

\subsection{Inference using Nearest Neighbor Search (NNS)} \label{sec:inference_optim_method}
As described in Appendix \ref{app:inference}, while making a prediction in SLDA, $\mu$ and $\Sigma$ are translated to $W$ and $b$ of an equivalent fully connected layer.
% During prediction in an SLDA classifier, the $\mu$ and $\Sigma$ matrices of SLDA are translated to $W$ and $b$ (weights and bias)
% of a fully connected layer equivalent, as described in \citet{hayes2020lifelong} and included in Appendix \ref{app:SLDA}.
The logits for all N classes are computed as per $y = W x^{T} + b$ where x denotes the embedding of input images.
Since this then requires the expensive normalization of logits against all classes, we optimize inference for sub-linear search using NNS.

Inspired by k-NN Softmax \cite{song2020knnsoftmax} and ANN-Softmax \cite{zhao2021annsoftmax}
% and their use of nearest neighbor algorithms for \emph{active class selection}
, we propose the use of Locality Sensitive Hashing (LSH) for \emph{active class selection} in XLDA.
% In the case of a fully connected layer, the nearest neighbor optimization picks out the k-nearest weight vectors $w_{k}$ from $W$ before predicting logits for these active classes, 
% In the case of a fully connected layer, the nearest neighbor optimization picks out the k-nearest weight vectors in the search index $\tilde{w}$ from $W$ before predicting logits for these active classes,
% and setting the logits of all other classes to zero.
% In our case, we optimize XLDA for inputs $x$ by searching for the \emph{k-nearest mean vectors} $\mu_{k}$ from $\mu$. We translate only the means of the active classes 
We optimize XLDA for inputs $x$ by searching for the \emph{k-nearest mean vectors} $\tilde{\mu}$ from all mean vectors $\mu$ present in the search index. We translate only the means of the active classes
% to the $w_{k}$ and $b_{k}$ to obtain the reduced fully connected layer equivalent. The classification logits are then computed as 
to $\tilde{w}$ and $\tilde{b}$ to obtain the reduced fully connected layer equivalent. The active classification logits are then computed as
$\tilde{y} =  \tilde{w} x^{T} +  \tilde{b}$ and the final classification logits as $y_i = \tilde{y}_i  \forall i \in$ k-active classes, and
$0$ for all others.

% \begin{equation}
% y_{k} = w_{k} x^{T} + b_{k}
% \tilde{y} =  \tilde{w} x^{T} +  \tilde{b}
% \end{equation}

% \begin{equation}
%   y =
%   \begin{cases}
%     % y_{k_{i}} & \text{if }  i \in \text{active classes} \\ 
%     \tilde{y}_{i} & \text{if }  i \in \text{k-active classes}    \\
%     0             & \text{if }  i \notin \text{k-active classes} \\
%   \end{cases}
% \end{equation}

In essence, this is a two step process of first running a \emph{nearest mean classifier} to shortlist the potential classes and then predicting on only those classes.

% \begin{table}
%   \centering
%   \caption{Inference Optimization}
%   \begin{tabular}{c|c|c}
%     \hline
%     Dataset     & SLDA  & XLDA           \\
%     \hline
%     AliProducts & 10 ms & \textbf{8 ms}  \\
%     Landmarks   & 16 ms & \textbf{10 ms} \\
%     \hline
%   \end{tabular}
% \end{table}

\begin{table*}[t]
  \centering
  \caption{Accuracies of Base Initialization}
  \label{tab:base_init}
  \begin{tabular}{c|c|c|c|c}
    \hline
    \textbf{Model Initialization}           & \textbf{CalTech Birds}   & \textbf{ImageNet-1k}     & \textbf{AliProducts}     & \textbf{Landmarks (Clean)} \\
                                            & (200 classes)            & (1000 classes)           & ($\sim$50k classes)      & ($\sim$80k classes)        \\
    \hline
    Dense Layer (Roof-line)                 & 72.07\%                  & 83.86\%                  & 81.35\%                  & 72.57\%                    \\
    \hline
    $\mu = W$ \& $\Sigma = I$               & \textbf{71.49\%} (-0.58) & \textbf{83.82\%} (-0.04) & 77.62\% (-3.73)          & 63.45\% (-9.12)            \\
    % \hline
    $\mu = W$ \& $\Sigma = \mathrm{Cov}(W)$ & 71.13\% (-0.94)          & 82.26\% (-1.6)           & \textbf{81.77\%} (+0.42) & \textbf{66.71\%} (-5.86)   \\
    \hline
  \end{tabular}
\end{table*}

% \begin{table*}[t]
%   \centering
%   \caption{SLDA Training Times}
%   \label{tab:parallel}
%   \begin{tabular}{c|c|c|c|c|c}
%     \hline
%     \textbf{Dataset}  & \textbf{\# Samples} & \textbf{\# Classes} & \textbf{FC Layer} & \textbf{SLDA}  (bs=1) & \textbf{XLDA} (bs=512) \\
%     % &                     &                     &                   & (batch size = 1) & (batch size = 512) \\
%     \hline
%     Caltech-Birds2011 & 6k                  & 200                 & 5s                & 7s                    & \textbf{0.36s}         \\
%     ImageNet-1k       & 1.3m                & 1000                & 33m               & 23m                   & \textbf{8s}            \\
%     iNaturalist 2018  & 0.45m               & 8142                & 1h15m             & 9m                    & \textbf{4s}            \\
%     AliProducts       & 2.5m                & 50000               & 44h               & 1h5m                  & \textbf{33s}           \\
%     \hline
%   \end{tabular}
% \end{table*}

\begin{table}[t]
  \centering
  \caption{Training time for 1 epoch. Speed-up factor of XLDA against FC layer shown in parentheses.}

  \label{tab:parallel}
  \begin{tabular}{c|c|c|c}
    \hline
    \textbf{Dataset} & \textbf{FC Layer} & \textbf{SLDA} & \textbf{XLDA}              \\
                     & (bs = 512)        & (bs = 1)      & (bs = 512)                 \\
    \hline
    Caltech-Birds    & \textbf{0.18s}    & 7s            & 0.36s (0.5$\times$)        \\
    ImageNet-1k      & 15s               & 23m           & \textbf{11s} (1.3$\times$) \\
    iNaturalist      & 37s               & 9m            & \textbf{5s} (7.4$\times$)  \\
    AliProducts      & 26m               & 1h5m          & \textbf{36s} (42$\times$)  \\
    \hline
  \end{tabular}
\end{table}

\section{Equivalence of LDA classifier and Fully Connected layer} \label{sec:equivalence}

A fully-connected layer with weights $W$ and bias $b$, linearly transforms the input $x$, resulting in a transformed output $Wx^T + b$. When activated with a softmax, this takes the form of the posterior $\hat{y} = p(y|x;W,b)$. Training classifiers with fully-connected layers, is effectively, learning the posterior distribution.

LDA classifier, on the other hand, models the likelihood over $x$, assuming normally-distributed samples $x \sim \mathcal{N}(\mu^{(y)}, \Sigma_s)$, where $\mu^{y}$ refers to the mean vector belonging to class $y$. Covariance $\Sigma_s$ is shared across classes for multivariate data, which is imperative for a linear decision boundary. Labels $y \in \mathcal{Y}$ are assumed to be multinomially distributed $y \sim \mathrm{Multinomial}(\phi^{(y)})$. In classification, we're interested in the posterior $p(y|x)$, we arrive at the posterior for LDA using Bayes' rule as shown in Eq.\refeq{eq:bayes_rule}.
\begin{align} \label{eq:bayes_rule}
  p(y|x)   & = \frac{p(x|y) p(y)}{p(x)}                                \\
  p(y=1|x) & = \frac{p(x|y=1) p(y=1)}{\Sigma_{j=0}^{1} p(x|y=j)p(y=j)}
\end{align}
For simplicity, we take the binary classification case with univariate inputs and shared variance $\sigma_s$. Prior $p(y=1)$ now becomes $y \sim \mathrm{Bern}(\phi)$, where $\phi$ is the probability of sampling $y=1$ in a single trial.
\begin{align}\label{eq:univariate_normal}
  \mathcal{N}(\mu,\sigma^2) & = \frac{1}{\sigma \sqrt{2\pi} } e^{-\frac{1}{2}\left(\frac{x-\mu}{\sigma}\right)^2} \\
  \mathrm{Bern(\phi^{(k)})} & = \begin{cases}
    \phi   & \text{if }k=1,     \\
    1-\phi & \text {if } k = 0.
  \end{cases}
\end{align}
Substituting distributions in Eq.\refeq{eq:bayes_rule} assuming binary outcomes, the posterior $p(y|x)$ simplifies to a logistic form, as shown below:
\begin{align} \label{eq:posterior_form}
  p(y=1|x) & = \frac{1}{1 + \frac{p(x|y=0)p(y=0)}{p(x|y=1)p(y=1)}}                                                                   \\
           & = \frac{1}{1 + \frac{\phi}{1-\phi} \frac{\exp(\frac{-(x-\mu_0)^2} {2\sigma^2})}{\exp (\frac{-(x-\mu_1)^2}{2\sigma^2})}} \\
           & = \frac{1}{1 + \alpha e^{(\frac{-1}{2\sigma^2} [\mu_0^2 - \mu_1^2 -2x(\mu_0 - \mu_1)])}}                                \\
           & \approx \frac{1}{1 + \alpha e^{-(wx + b)}}
\end{align}
Here $\alpha = \frac{\phi}{1-\phi}$. Assuming equal prior, $\alpha$ is eliminated, and the resulting expression now resembles a logistic curve with parameters $w$ and $b$.

Hence, an FC layer with binary outcome essentially learns the posterior $p(y|x)$, that resembles the logistic function with parameters $W$ and $b$, which is implicitly a function of $\mu^{(y)}$ and shared variance $\sigma_s$.

For the sake of brevity, we show the equivalent reductions for $W$ and $b$ for class $k$ in case of multivariate input $\textbf{x}$ below. More details can be found in Appendix \ref{app:SLDA}.
\begin{align}\label{eq:wb_from_mu_sigma}
  \hat{\textbf{y}} & = \mathrm{W}^T\textbf{x} + \mathrm{b}                     \\
  \Lambda          & =((1-\beta)\Sigma+\beta\mathrm{I})^{-1}, \beta \in (0, 1) \\
  \mathrm{w}_k     & = \Lambda \mu_k, k \in \mathcal{Y}                        \\
  \mathrm{b}_k     & = -\frac{1}{2}(\mu_k .\Lambda \mu_k)
\end{align}
Therefore, given parameters $W$ and $b$ from an FC layer, assuming normally-distributed data and shared identity covariance, one can derive `initial' guesses for $\mu=W$ and $\Sigma_s=\mathrm{I}$, to enable CL without retraining LDA. Experiments on these initializations are discussed in Section \ref{sec:conversion}.

Conversely, an LDA layer can be converted to an FC layer for end-to-end training using $W$ and $b$ initialized based on the above parameters. As we show in Section \ref{sec:conversion}, this provides compute savings up to \textbf{42$\times$} for up to 50k classes, at a minimum, since SLDA converges to an optimum starting point of $W$ and $b$ in a \textbf{single} epoch, where an FC layer typically requires multiple epochs to converge on a solution. This speedup gets significant, when accounting for multiple epochs that FC layers require.

% \begin{table}
%   \centering
%   \caption{Results for Nearest Neighbor (NN) Search Inference Optimization in XLDA}
%   \label{tab:inference}
%   \begin{tabular}{|c|c|c|}
%     \hline
%     \textbf{Dataset}  & \textbf{SLDA} & \textbf{XLDA} \\
%     \hline
%     CalTech Birds     & 71.49         & 83.82         \\
%     \hline
%     ImageNet1k        & 73.20         & 83.64         \\
%     \hline
%     AliProducts       & 73.32         & 81.50         \\
%     \hline
%     Landmarks (Clean) & 72.18         & 82.27         \\
%     \hline
%   \end{tabular}
% \end{table}

% \begin{table}
%   \centering
%   \caption{Results}
%   \label{tab:inference}
%   \begin{tabular}{|c|c|c|}
%     \hline
%     \textbf{Dataset}  & \textbf{SLDA} & \textbf{XLDA} \\
%     \hline
%     AliProducts       & 1             & 2             \\
%     \hline
%     Landmarks (Clean) & 16            & 10            \\
%     \hline
%   \end{tabular}
% \end{table}

\section{Experiments and Results}\label{sec:experiments}
Details of the experimental setup and the data used are included in Appendices \ref{app:exp_setup} and \ref{app:validation_sets}.

% We primarily use the datasets in Table \ref{tab:datasets} for our experiments on
% extreme classification. Additional datasets are used to demonstrate the results of
% individual optimizations of XLDA.
% All experiments were designed in Tensorflow and run on (Add specs of machine here).
% \begin{itemize}
%   \item \textbf{Aliproducts} \cite{le2020sine_aliprod} 
%   \item \textbf{Google Landmarks} \cite{weyand2020landmarks}
%   \item \textbf{Met dataset} \cite{ypsilantis2021met}
% \end{itemize}
% Maybe make this a table with all the dataset statistics?

% \setcitestyle{authoryear,round,citesep={;},aysep={,},yysep={;}}

\subsection{Conversion from a Fully-Connected Layer}\label{sec:conversion}
% We present our results on initializing SLDA with the weights of fully connected layer
% as described in Sec. \ref{sec:base_init}
% in Table \ref{tab:base_init}.
% The CNN backbone used for our experiments is an EfficientNetv2-S (Add citation).
We train a Fully Connected (Dense) layer classifier with the EfficientNetV2-S \cite{tan2021efficientnetv2} backbone and
then attempt to transfer the parameters of the Dense layer to a CL-capable SLDA.
% by simply initializing with the weights of the Dense layer.
For ImageNet-1k we use the pretrained classifier weights of EfficientNet-V2-S for initialization.
In Table \ref{tab:base_init}, we compare the accuracies of the initialized SLDA model with the Dense layer that was used
to initialize it. Validation sets used have been detailed in Appendix \ref{app:validation_sets}.
% instead of training the
% Dense classifier, as done for other datasets.

We notice that a major portion of the accuracy from the Dense layer is retained in the
SLDA classifier and the initialization only results in a small drop-off in accuracy which can be considered
a trade-off for CL capabilities.
Between the two options for initializing $\Sigma$ --- $I$ and $\mathrm{Cov}(W)$ --- $I$ seems to be the better option
for datasets with a smaller number of classes due to lower inter-class confusion and in turn, lower correlation between the dimensions of embeddings.
$\mathrm{Cov}(W)$ seems to be the better option for datasets with a large number of fine-grained classes due to higher inter-class confusion
which breaks the assumption of uncorrelated embedding space.
%  Hence, the usage of the $\mathrm{Cov}(W)$ approximation
% helps counter the confusion and correlation.

The results for ImageNet-1k also demonstrate the capability to transfer pre-trained weights of
a Dense classifier head trained without a CL objective or architecture, to a CL-amenable and edge deployment-ready SLDA classifier straightaway
with almost no drop in accuracy.

We study the improvement by training $\mu$ and $\Sigma$, and the impact of each component in Appendix \ref{app:ablation}.

% \begin{table*}
%   \centering
%   \caption{Results of Base Initialization}
%   \label{tab:base_init}
%   \begin{tabular}{|c|c|c|c|}
%     \hline
%     \textbf{Model Initialization}                & \textbf{CalTech Birds}     & \textbf{AliProducts}     & \textbf{ImageNet1k}      \\
%     \hline
%     Dense Layer                                  & 72.07\%                    & 81.35\%                  & 83.86\%                  \\
%     \hline
%     \hline
%     $\mu = W$ and $\Sigma = I$                   & \textbf{71.49\%} (-0.58\%) & 77.62\% (-3.73)          & \textbf{83.82\%} (-0.04) \\
%     \hline
%     $\mu = W$ and $\Sigma = Cov(W)$              & 71.13\% (-0.94\%)          & \textbf{81.77\%} (+0.42) & 82.26\% (-1.6)           \\
%     \hline
%     \hline
%     $\mu$ fit from scratch and $\Sigma = I$      & 73.20\% (+1.71)            & 82.66\%  (+0.89)         & 83.64\% (-0.18)          \\
%     \hline
%     $\mu$ fit from scratch and $\Sigma = Cov(W)$ & 73.32\% (+1.83)            & 84.65\% (+2.88)          & 81.50\%  (-2.32)         \\
%     \hline
%     $\mu = W$ and $\Sigma$ fit from scratch      & 72.18\% (+0.69)            & 78.76\%  (-3.01)         & 82.27\% (-1.55)          \\
%     \hline
%     $\mu$ and $\Sigma$ both fit from scratch     & 74.90\%  (+3.41)           & 86.87\% (+5.1)           & 83.74\% (-0.08)          \\
%     \hline
%   \end{tabular}
% \end{table*}

\subsection{Batch Parallelization}

On datasets of increasing sample size and class count, the results in Table \ref{tab:parallel} compare the time taken to train SLDA sample-wise and batch-wise, with the time to train a fully
connected layer with the dataset embeddings as reference. The FC layer is trained for 1 epoch with cosine decay and a batch size of 512. However, 1 epoch is not sufficient
to train an FC layer, in which case the speedup for XLDA is even more pronounced (as shown in Table \ref{tab:parallel_100ep}).
% SLDA trained with samples in succession showcases the theoretical $O(1)$ training complexity in the number of classes,
% as opposed to an FC layer's $O(M)$. 
% The training times for the SLDA classifier trained with samples in succession only increase relative to the number of samples in the
% dataset and not the class count.
By implementing vectorized SLDA over batches of 512 samples, training times significantly decrease to less than a minute,
even for the large-scale AliProducts datasets of around 2.5M samples. XLDA gains exponentially versus FC as the number of classes grow, due to its $O(1)$ training construction.

\subsection{Inference optimization}\label{sec:inference_opt}
\begin{table}
  \centering
  \caption{Per-sample Inference times and Speed-up factors}
  \label{tab:inference}
  \begin{tabular}{c|c|c}
    \hline
    \textbf{Dataset} & \textbf{SLDA} & \textbf{XLDA (Ours)}         \\
    \hline
    ImageNet-1k      & 7 ms          & \textbf{5 ms} (1.4$\times$)  \\
    AliProducts      & 172 ms        & \textbf{41 ms} (4.2$\times$) \\
    Landmarks        & 278 ms        & \textbf{59 ms} (4.7$\times$) \\
    \hline
  \end{tabular}
\end{table}

We evaluate the performance of Nearest Neighbor (NN) Search optimization in XLDA
by comparing the time taken to make predictions per-sample (batch size = 1) during inference in Table \ref{tab:inference}.
We report times for inference done purely on CPU; both the LSH Nearest Neighbor Search and the logit computation.
We choose the number of active classes $k = \frac{C}{10}$ where $C$ is the total number of classes.
We notice that XLDA is faster per-sample, and over the entire test set we observe upto a
$\sim$ 5$\times$ speedup. The speedup is more pronounced on datasets with a larger number of classes.

\section{Conclusion and Future Work}
In this paper, we've discussed XLDA, a framework to extend Class-IL to classification on a large number
of classes, and described a suite of optimizations to efficiently train and
use it for inference at compute constrained edge devices.
Our results for converting an FC layer to an LDA classifier with simple
initializations are promising and in future works,
a conversion that enables end-to-end training for LDA classifiers could be explored.

\bibliography{refs}
\bibliographystyle{icml2023/icml2023}

%%%%%%%%%%%%%%%%%%%%%%%%%%%%%%%%%%%%%%%%%%%%%%%%%%%%%%%%%%%%%%%%%%%%%%%%%%%%%%%
%%%%%%%%%%%%%%%%%%%%%%%%%%%%%%%%%%%%%%%%%%%%%%%%%%%%%%%%%%%%%%%%%%%%%%%%%%%%%%%
% APPENDIX
%%%%%%%%%%%%%%%%%%%%%%%%%%%%%%%%%%%%%%%%%%%%%%%%%%%%%%%%%%%%%%%%%%%%%%%%%%%%%%%
%%%%%%%%%%%%%%%%%%%%%%%%%%%%%%%%%%%%%%%%%%%%%%%%%%%%%%%%%%%%%%%%%%%%%%%%%%%%%%%
\newpage
\appendix
\onecolumn

\section{Ablation Study for FC-XLDA conversion} \label{app:ablation}
In Table \ref{tab:base_ablation}, we study the improvement by training the two components of SLDA --- $\mu$ and $\Sigma$ ---
as opposed to initializing it from $W$, as well as the relative impact of training each of these components on accuracy. We notice that training only $\mu$  leads to a more significant
boost in performance compared to training only $\Sigma$. As expected training both $\mu$ and $\Sigma$  leads
to a significant boost in performance, albeit at the cost of training time and compute.

\begin{table*}[h!]
  \centering
  \caption{Ablation study for improvement by training each component of SLDA over initializing SLDA}
  \label{tab:base_ablation}
  \begin{tabular}{c|c|c|c}
    \hline
    \textbf{SLDA Model Initialization}                    & \textbf{CalTech Birds}    & \textbf{ImageNet-1k}     & \textbf{AliProducts}    \\
    \hline
    $\mu = W$ \& $\Sigma = I/\mathrm{Cov}(W)$ (Best Init) & 71.49\%                   & 83.82\%                  & 81.77\%                 \\
    \hline
    $\mu$ trained  \& $\Sigma = I$                        & 73.20\% (+1.71)           & 83.64\% (-0.18)          & 82.66\%  (+0.89)        \\
    % \hline
    $\mu$ trained  \& $\Sigma = \mathrm{Cov}(W)$          & 73.32\% (+1.83)           & 81.50\%  (-2.32)         & 84.65\% (+2.88)         \\
    % \hline
    $\mu = W$ \& $\Sigma$ trained                         & 72.18\% (+0.69)           & 82.27\% (-1.55)          & 78.76\%  (-3.01)        \\
    % \hline
    $\mu$ \& $\Sigma$ both trained                        & \textbf{74.90\%}  (+3.41) & \textbf{83.74\%} (-0.08) & \textbf{86.87\%} (+5.1) \\
    \hline
  \end{tabular}
\end{table*}

\section{Background on SLDA} \label{app:SLDA}

Streaming Linear Discriminant Analysis (SLDA) was introduced by \citet{hayes2020lifelong} as a
linear classifier that makes a prediction based on embeddings from a fixed CNN-backbone.
The SLDA classifier stores a mean vector $\mu_k$ per class, which together make the mean matrix $\mu$, as well
as a shared covariance matrix $\Sigma$. This allows SLDA to learn a per-class distribution, giving it generative properties
that are ideal for Class-IL.

\subsection{Training} \label{app:SLDA_train}
During training, in addition to $\mu_k$, an associated count of samples seen for the class $k$ is stored and is represented by $c_k$.
On the arrival of a new sample $(\mathbf{z}_t, y)$, the mean $\boldsymbol{\mu}_{y}$ parameter and counter variable $c_{y}$
of the class at a timestep $t$ are updated by

\begin{equation}
  \begin{aligned}
    \boldsymbol{\mu}_{(k=y,t+1)} \gets \frac{c_{(k=y,t)}\boldsymbol{\mu}_{(k=y,t)} + \mathbf{z}_t}{ c_{(k=y,t)} + 1} \\
    c_{(k=y,t+1)} = c_{(k=y,t)} + 1
  \end{aligned}
\end{equation}

The shared covariance matrix $\Sigma$ is either kept plastic or fixed --- plastic refers to the
covariance being updated at every training step after base initialization and fixed is when it does not change
after base-initialization. In the plastic scenario, $\Sigma$ is updated as per
\begin{equation}
  \begin{aligned}
    \Sigma_{t+1}  = \frac{t\Sigma_t + \Delta_t}{t+1} \\
  \end{aligned}
\end{equation}

\begin{equation}
  \begin{aligned}
    \Delta_t     = \frac{t(z_t - \mu_{(k=y,t)})(z_t - \mu_{(k=y,t)})^T}{t+1}
  \end{aligned}
\end{equation}

\subsection{Inference} \label{app:inference}
To use SLDA as a classifier and make predictions, $\mu$ and $\Sigma$ are converted to an equivalent
fully-connected layer $W$ and $b$ as per
\begin{equation}
  w_k = \Lambda \mu_k
\end{equation}

\begin{equation}
  b_k = - \frac{1}{2} (\mu_k \cdot \Lambda \mu_k)
\end{equation}

where $\Lambda \sim \Sigma^{-1}$ with a shrinkage estimator applied. This implies,
\begin{equation} \label{eq:mu_w_flip}
  \begin{aligned}
    W = \Sigma^{-1} \mu \\
    \mu = \Sigma W
  \end{aligned}
\end{equation}

\begin{table}[h!]
  \centering
  \caption{Dataset Statistics of number of classes and training samples}
  \label{tab:datasets}
  \begin{tabular}{c|c|c}
    \hline
    \textbf{Dataset}                                                                         & \textbf{\# classes} & \textbf{\# samples} \\
    \hline
    \textbf{CalTech Birds (2011)} \cite{WelinderEtal2010}                                    & 200                 & 6k                  \\
    \textbf{ImageNet-1K (2012)} \cite{ILSVRC15}                                              & 1K                  & 1.3m                \\
    % \hline
    \textbf{i-Naturalist} \cite{van2018inaturalist}                                          & 8.1K                & 0.5m                \\
    \textbf{AliProducts v1} \cite{le2020sine_aliprod}                                        & 50K                 & 2.5m                \\
    % \hline
    \textbf{Google Landmarks v2 - Clean} \cite{weyand2020landmarks, yokoo2020landmarksclean} & 81K                 & 1.4m                \\
    \hline
  \end{tabular}
\end{table}

\section{Experimental Setup} \label{app:exp_setup}
Our experiments were implemented in TensorFlow and run on a Intel Core i9-10920X CPU
with a NVIDIA RTX A6000 GPU.

\section{Data and Validation Sets} \label{app:validation_sets}
We use the datasets shows in Table \ref{tab:datasets} for our experiments.
Google Landmarks v2 \cite{weyand2020landmarks} has 200k classes originally, but we use the
Clean version of the dataset created by \citet{yokoo2020landmarksclean}, as it is more amenable to
classification tasks, compared to the original which is better suited for retrieval. The clean version consists of classes with at least 2 samples, reducing the class count to 81k.

The validation set used for reporting accuracy were created as described below.
For CalTech Birds, ImageNet-1k and i-Naturalist we use the train test split
created as part of the Tensorflow-Hub dataset.
For AliProducts we use the train and validation splits created
for the AliProducts Product Recognition Challenge 2020, as is. For Google Landmarks (Clean), we create a 1-sample per class validation set, i.e
our validation set consists of 81k images which correspond to the 81k classes in the dataset.

\begin{table}[t]
  \centering
  \caption{Practical training times for FC layer (as opposed to ideal 1 epoch). Speedup with XLDA training for 1 epoch}

  \label{tab:parallel_100ep}
  \begin{tabular}{c|c|c|c}
    \hline
    \textbf{Dataset} & \textbf{Approx. \# epochs}  & \textbf{FC Layer} & \textbf{Practical Speedup} \\
                     & \textbf{for FC to converge} & (bs = 512)        & \textbf{of XLDA}           \\
    \hline
    Caltech-Birds    & 25                          & 4.5s              & 12.5$\times$               \\
    iNaturalist      & 20                          & 740s              & 148$\times$                \\
    AliProducts      & 15                          & 390m              & 650$\times$                \\
    \hline
  \end{tabular}
\end{table}

%%%%%%%%%%%%%%%%%%%%%%%%%%%%%%%%%%%%%%%%%%%%%%%%%%%%%%%%%%%%%%%%%%%%%%%%%%%%%%%
%%%%%%%%%%%%%%%%%%%%%%%%%%%%%%%%%%%%%%%%%%%%%%%%%%%%%%%%%%%%%%%%%%%%%%%%%%%%%%%

\end{document}